\documentclass[sigconf]{acmart}
\renewcommand\footnotetextcopyrightpermission[1]{} 
\acmConference{}{}{} 
\acmYear{}
\acmISBN{}
\acmDOI{}

\AtBeginDocument{%
  \providecommand\BibTeX{{%
    \normalfont B\kern-0.5em{\scshape i\kern-0.25em b}\kern-0.8em\TeX}}}

\usepackage{enumitem}
\usepackage{multirow}
\usepackage{multicol}
\usepackage{booktabs}
\usepackage{colortbl} 
\usepackage{amsmath}
\usepackage{amssymb}
\usepackage{stmaryrd}
\usepackage{pifont}
\usepackage{subcaption}
\usepackage{diagbox}
\begin{document}
\setlength{\floatsep}{2pt plus 2pt minus 1pt}
\setlength{\textfloatsep}{2pt plus 2pt minus 1pt}
\setlength{\intextsep}{2pt plus 2pt minus 1pt}
\title{D2SP: Dynamic Dual-Stage Purification Framework for Dual Noise Mitigation in Vision-based Affective Recognition.}
\author{Haoran Wang}
\author{Xinji Mai}
\author{Zeng Tao}

\affiliation{%
  \institution{Shanghai Engineering Research Center of AI \& Robotics, Academy for Engineering \& Technology, Fudan University, Shanghai, China.}
  \city{Shanghai}
  \country{China}
}
\email{hrwang23@m.fudan.edu.cn}

\author{Xuan Tong}
\author{Junxiong Lin}
\author{Yan Wang}
\authornote{Corresponding Author}
\affiliation{%
  \institution{Shanghai Engineering Research Center of AI \& Robotics, Academy for Engineering \& Technology, Fudan University, Shanghai, China.}
  \city{Shanghai}
  \country{China}
}
\email{yanwang19@fudan.edu.cn}

\author{Jiawen Yu}
\author{Boyang Wang}
\author{Shaoqi Yan}
\author{Qing Zhao}
\affiliation{%
  \institution{Shanghai Engineering Research Center of AI \& Robotics, Academy for Engineering \& Technology, Fudan University, Shanghai, China.}
  \city{Shanghai}
  \country{China}
}

\author{Ziheng Zhou}
\affiliation{%
  \institution{School of Information Science and Technology, Fudan University}
  \city{Shanghai}
  \country{China}
}

\author{Shuyong Gao}
\affiliation{%
  \institution{Fudan University}
  \city{Shanghai}
  \country{China}
}

\author{Wenqiang Zhang}
\authornotemark[1]
\affiliation{%
  \institution{Engineering Research Center of AI \& Robotics, Ministry of Education, Academy for Engineering \& Technology, Fudan University. \and
Shanghai Key Lab of Intelligent Information Processing, School of Computer Science, Fudan University}
  \city{Shanghai}
  \country{China}
}
\email{wqzhang@fudan.edu.cn}

%



\begin{abstract}

The contemporary state-of-the-art of Dynamic Facial Expression Recognition (DFER) technology facilitates remarkable progress by deriving emotional mappings of facial expressions from video content, underpinned by training on voluminous datasets. Yet, the DFER datasets encompass a substantial volume of noise data. Noise arises from low-quality captures that defy logical labeling, and instances that suffer from mislabeling due to annotation bias, engendering two principal types of uncertainty: the uncertainty regarding data usability and the uncertainty concerning label reliability. Addressing the two types of uncertainty, we have meticulously crafted a two-stage framework aiming at \textbf{S}eeking \textbf{C}ertain data \textbf{I}n extensive \textbf{U}ncertain data (SCIU). This initiative aims to purge the DFER datasets of these uncertainties, thereby ensuring that only clean, verified data is employed in training processes. To mitigate the issue of low-quality samples, we introduce the Coarse-Grained Pruning (CGP) stage, which assesses sample weights and prunes those deemed unusable due to their low weight. For samples with incorrect annotations, the Fine-Grained Correction (FGC) stage evaluates prediction stability to rectify mislabeled data. Moreover, SCIU is conceived as a universally compatible, plug-and-play framework, tailored to integrate seamlessly with prevailing DFER methodologies. Rigorous experiments across prevalent DFER datasets and against numerous benchmark methods substantiates SCIU's capacity to markedly elevate performance metrics.

\end{abstract}


\begin{teaserfigure}
\centering
  \includegraphics[width=0.8\textwidth]{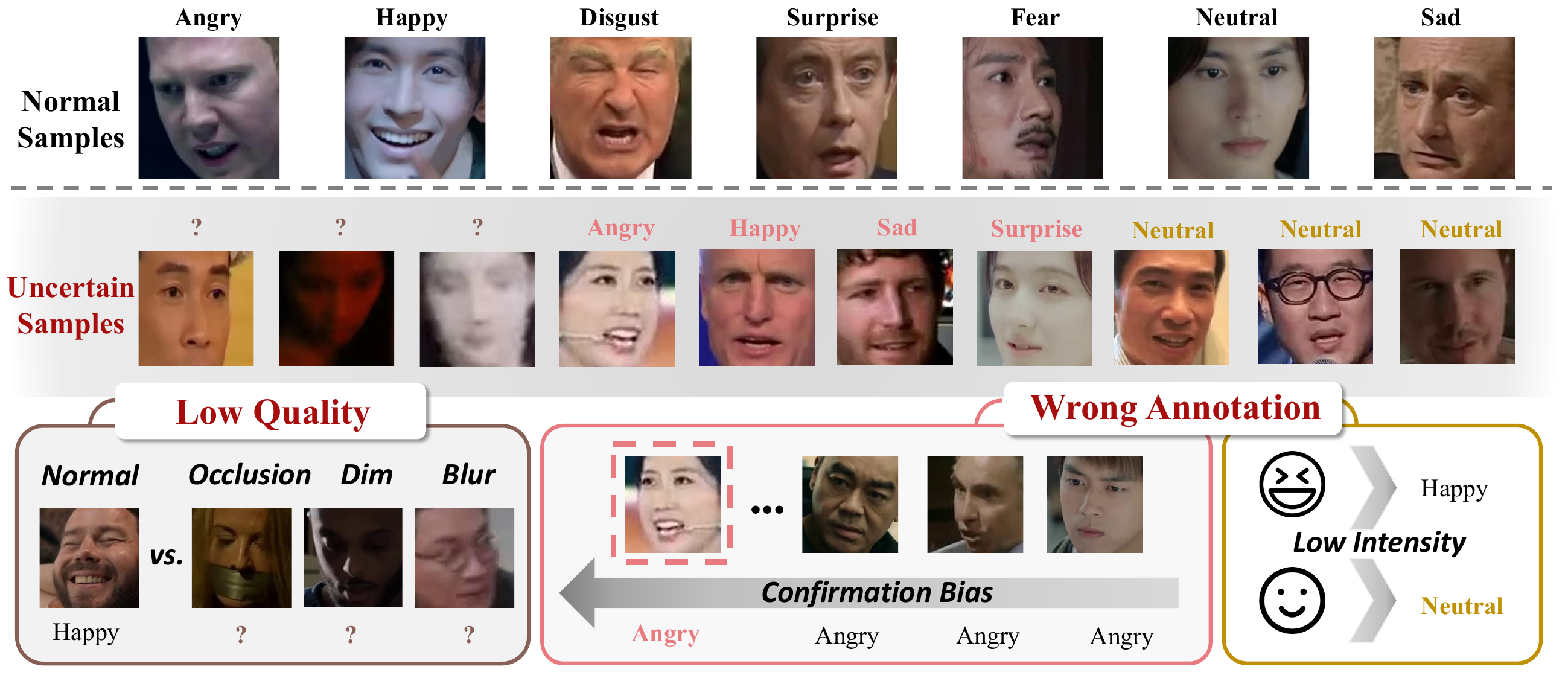}
  \caption{Illustration of a selection of noisy samples from the FERV39k and DFEW datasets, representing the two principal types of uncertain samples: low quality and wrong annotation. The low quality of samples introduce uncertainty regarding their usability and the wrong annotation of samples introduce uncertainty about the accuracy of their labels. Low-quality samples are attributable to factors such as occlusion, dim lighting, blurriness, etc.  Wrong-annotation samples  result from reasons such as: (1) the confirmation bias of annotators leading to misclassificaiton; (2) low-intensity of emotion, which are often incorrectly labeled as Neutral.}
  \Description{}
  \label{fig1}
\end{teaserfigure}
\maketitle
\section{Introduction}
Facial Expression Recognition (FER) is essential in sectors such as Human-Robot Interaction (HRI) \cite{HMI10018501} and healthcare, with Dynamic Facial Expression Recognition (DFER) emphasizing the importance of temporal information to improve the accuracy of FER. Recently, the field of DFER tasks has developed rapidly, thanks to increasingly powerful models. Since the adoption of traditional Convolutional Neural Networks (CNNs) like ResNet \cite{he2016deep} for singular video frame feature extraction, through to the incorporation of Recurrent Neural Networks (RNNs)  \cite{singh2020facial} or Long Short-Term Memory (LSTM) \cite{li2019cnn} networks for gleaning temporal information, and the application of Vision Transformers (ViT)  \cite{dosovitskiy2020image, yuan2021tokens} as foundational architectures, the performance of methods has been heavily reliant on the evolution of increasingly sophisticated models to bolster feature detection capabilities \cite{DBLP:conf/mm/MorotoY0KSTMOH23, DBLP:conf/mm/XingTHL022, DBLP:conf/mm/ShaoWLHP022}. Techniques such as Freq-HD \cite{10.1145/3581783.3611972} and NR-DFERNet \cite{li2022nr} have highlighted the significance of choosing frames rich in expressive content to efficiently filter through video segments, capturing the nuances of facial expression transitions with enhanced accuracy. Nevertheless, these advancements often overlook the intrinsic complications posed by DFER datasets \cite{DBLP:conf/mm/WuSCJ23, DBLP:conf/mm/LiuCLLX23}.

Recent research efforts have shifted from controlled laboratory environments \cite{lucey2010extended,zhao2011facial} towards creating datasets in in-the-wild conditions \cite{kossaifi2017afew,jiang2020dfew,wang2022ferv39k}, which are more reflective of naturalistic environments. These datasets present a higher level of challenge due to the variability in expression intensity and the significant intra-class variation, leading to increased difficulty in DFER task, as well as bringing the low-quality and annotation bias issues. 
In FER task, uncertainty refers to a concept of classification ambiguity \cite{zheng2023facial, wang2022rethinking, DBLP:conf/mm/ZhuLSTZ023}, and extending such uncertainty to DFER datasets pertains to the ambiguity in annotations, which stems from both subjective and objective sources \cite{DBLP:conf/mm/0020PZYLWW23, DBLP:conf/mm/WuWC23}.
Based on our experiences during constructing dataset, along with extensive research and surveys \cite{wang2022ferv39k,jiang2020dfew,wang2022rethinking}, we have identified several factors contributing to uncertainty in DFER datasets. 
Through detailed scrutiny of prevailing DFER datasets, we discovered and defined two types of uncertainty: uncertainty of the usability of the data and uncertainty of the accuracy of the label. In Figure \ref{fig1}, we illustrate samples exhibiting these two types of uncertainties. The first type of high-uncertainty samples results from low-quality, rendered unusable due to issues like occlusion, dim lighting, blurriness, etc. This uncertainty arises from the objective issue of low-quality samples, which still need to be labeled for annotators rather than discarded. The second type of high-uncertainty samples stems from annotation bias, which originates from factors such as: confirmation bias leading to mislabelling, samples with emotional expressions being labeled as 'Neutral' due to low intensity of emotion, etc. Thus, this category of uncertainty is caused by the subjective factors of the annotators.

Here, we aim to further introduce the concept of confirmation bias. Confirmation Bias, derived from cognitive psychology, denotes the inclination of individuals to favor and recall information that corroborates their pre-existing beliefs, while overlooking or undervaluing evidence to the contrary \cite{nickerson1998confirmation}. This is notably evident in Semi-Supervised Learning (SSL), especially concerning the generation of pseudo-labels, where measures are taken to curtail or mitigate the effects of confirmation bias \cite{Arazo_Ortego_Albert_O’Connor_McGuinness_2020}. Typically, the discussion around confirmation bias centers on the production of pseudo-labels by teacher models \cite{Zhou_Li_2010,Xie_Luong_Hovy_Le_2019}. The generation of inaccurate pseudo-labels by these models can precipitate a feedback loop, where the model increasingly overfits to these incorrect labels, exacerbating the production of erroneous pseudo-labels \cite{li2023semireward,chen2023two,gilhuber2023overcome,ding2023combating,chen2022debiased}. In SSL, teacher networks are actually employed to annotate a large volume of unlabeled data. Analogously, when manually annotating, confirmation bias can emerge on annotators as well. This issue becomes particularly pronounced in datasets like DFER, where intra-class variance is substantial. For instance, if an annotator has consecutively labeled several samples with the same tag, they may subconsciously seek information confirming their prior labels when annotating subsequent samples. This can lead to the incorrect categorization of ambiguous or atypical expressions into these anticipated categories. We illustrate such a scenario in Figure \ref{fig1}.

This accumulation of bias during the data collection and annotation stages may culminate in a dataset that disproportionately reflects the preconceptions of annotators rather than the true diversity and complexity of expressions. Commonly, the process of building such datasets involves having multiple annotators work in batches \cite{jiang2020dfew, wang2022ferv39k}, inherently introducing varying degrees of confirmation bias. This bias, when present in the annotation of DFER datasets, leads to the creation of data that is inaccurately labeled. The presence of erroneous labels not only undermines the precision of these datasets but also predisposes the derived models towards perpetuating these inaccuracies, further amplifying the effects of confirmation bias.

To mitigate the prevalent uncertainty issues identified in DFER datasets, we introduce a pioneering dual-stage framework SCIU in DFER datasets, which is composed of Coarse-Grained Pruning (CGP) and Fine-Grained Correction (FGC). In the CGP stage, we prune the dataset and incorporate an innovative weighting branch within the network architecture, tasked with determining the weight of each sample, aiming to eliminate the first type of high-uncertainty samples, namely, low-quality, unusable samples. The dataset, post-first-stage pruning, is then advanced to the FGC stage, which is utilized to rectify mislabeled data. FGC focuses on correcting these inaccuracies by identifying samples with stable predictions in two dimensions, thereby allowing the second type of high-uncertainty samples—those with incorrect labels due to annotation bias—to be accurately relabeled.


Our proposed framework has facilitated notable enhancements in performance across various established DFER model architectures. The primary contributions of our research are outlined as follows:
    
    
\begin{itemize}
[leftmargin=*, noitemsep, topsep=0pt, partopsep=0pt]
    \item We identified two types of uncertainty within DFER datasets: uncertainty of the usability of the data and uncertainty of the accuracy of the label, which corresponding to two types of uncertain samples: 1) low-quality and unusable samples, 2) samples with wrong annotation from annotation biases such as confirmation bias of annotators and mislabelling low-emotion-intensity of samples as Neutral.
    
    \item We introduce a general and plug-and-play dual-stage framework SCIU, encompassing two stages: CGP for pruning the first type of uncertainty, and FGC aiming at correcting the second type of uncertainty in DFER datasets. 
    
    \item Through extensive experimentation on mainstream DFER datasets: FERV39k, DFEW and MAFW, and employing six prevalent backbone architectures within our SCIU framework, we demonstrated significant enhancements in performance. These results not only validate the existence of the uncertainty issues we proposed in DFER dataset but also attest to the general applicability and effectiveness of our framework.
\end{itemize}

\section{Related Work}


\subsection{Dynamic Facial Expression Recognition}


In the realm of deep learning, foundational work utilizing 3D Convolutional Neural Networks (C3D) \cite{tran2015learning} has been instrumental in the DFER domain, enabling the simultaneous capture of spatial and temporal information \cite{tran2015learning}. The subsequent integration of RNNs \cite{li2019cnn} and LSTMs \cite{singh2020facial} has markedly enhanced the ability to model temporal dynamics. These advancements have been complemented by traditional feature extraction techniques employing architectures such as ResNet \cite{he2016deep} and VGG \cite{simonyan2014very}. Methods such as Freq-HD \cite{10.1145/3581783.3611972} and NR-DFERNet \cite{li2022nr} focus on selecting frames with emotions. More recently, the adaptation of Transformer architectures from the NLP sector to the visual domain, through ViT \cite{dosovitskiy2020image}, has introduced a novel approach for feature extraction from images \cite{sun2023mae,chen2023static,sun2024hicmae,li2023intensity,lee2023frame}. And methods \cite{tao20243,li2023cliper,zhao2023prompting,foteinopoulou2023emoclip} based on CLIP-ViT have been proposed to address the DFER task due to CLIP’s powerful prior knowledge. The self-attention mechanism inherent in ViTs offers improved efficacy in discerning relevant information from images for DFER tasks. Moreover, the introduction of Former-DFER \cite{zhao2021former}, which utilizes both spatial and temporal transformers, has demonstrated commendable performance, further substantiating the potential of transformer-based models in addressing DFER challenges.

Diverging from the aforementioned approaches, our work hinges on eliminating the uncertainties prevalent within the DFER dataset. This ensures that the data learned by the model is accurate, reliable, and of high quality, thereby augmenting the model's recognition accuracy.

\vspace{-4mm}
\subsection{Learning with Uncertainty}
\textbf{Learning with Noisy Label. }Uncertainty refers to annotation ambiguity in FER task, which belongs to noisy label. The issue of learning with noisy label has garnered significant interest in the machine learning community in recent years \cite{DBLP:conf/mm/ZhangHBX22,DBLP:conf/mm/TanXWL21,feng2023rono,tu2023learning,ma2020normalized,liu2015classification}. This surge in attention is largely due to the recognition that noise is an inevitable aspect of the dataset labeling process \cite{DBLP:conf/mm/ZhangHBX22, DBLP:conf/mm/Li022}. Initial approaches were primarily focused on Sample Selection, a method for pruning data that contains noise \cite{feng2023ot,Bai_Yang_Han_Yang_Li_Mao_Niu_Liu_2021,Cheng_Liu_Ning_Wang_Han_Niu_Gao_Sugiyama_2022}. Techniques such as utilizing small-loss samples for training \cite{jiang2018mentornet,han2018co,Cheng_Liu_Ning_Wang_Han_Niu_Gao_Sugiyama_2022,Yu_Han_Yao_Niu_Tsang_Sugiyama_2019}—for instance, MentorNet \cite{jiang2018mentornet} employs the small loss principle for the teacher model to select clean samples for the student model. Alternatively, some strategies involve adopting robust loss functions to mitigate the learning of noise, such as Mean Absolute Error (MAE) loss \cite{ghosh2017robust} and Symmetric Cross Entropy (SCE) loss \cite{wang2019symmetric}. Methods like SELF \cite{Nguyen_Mummadi_Ngo_Nguyen_Beggel_Brox_2020} and SFT \cite{wei2022self} rely on predictions from previous iterations to discern clean samples.
\\ \textbf{Uncertainty in Facial Expression Recognition. }Given the unique nature of noise in FER, which encompasses both data quality issues and the subjectivity of human labeling, labels are imbued with considerable uncertainty \cite{wang2023rethinking,Zeng_Shan_Chen_2018}. To tackle this uncertainty, several innovative methods have been proposed. An intuitive approach involves measuring and learning the distribution of uncertainty, such as the DUL \cite{chang2020data}, which concurrently learns the distribution of features and uncertainty by utilizing the mean for features and variance for uncertainty, with the inference phase focusing solely on the mean to mitigate noise effects. Other strategies, such as the SCN \cite{wang2020suppressing}, operate by evaluating and classifying the significance of samples within a batch, assigning higher learning weights to those deemed of higher importance and relabeling select samples from the lower importance category based on specific criteria. This technique effectively minimizes the uncertainty prevalent in FER datasets. Additionally, methods like GAAVE \cite{zheng2023attack} assess sample uncertainty through sample perturbation, operating under the principle that certain samples exhibit robustness, whereas uncertain samples are more susceptible to disturbances.

Nevertheless, these methodologies primarily cater to Static Facial Expression Recognition (SFER), and DFER introduces additional complexities due to the temporal variability of samples and the multifaceted nature of the uncertainty involved. Thus, addressing the two types of uncertainty identified in DFER dataset needs more attention and is urgent to solve.

\begin{figure*}[htbp]
\centering
	\includegraphics[width=1\textwidth]{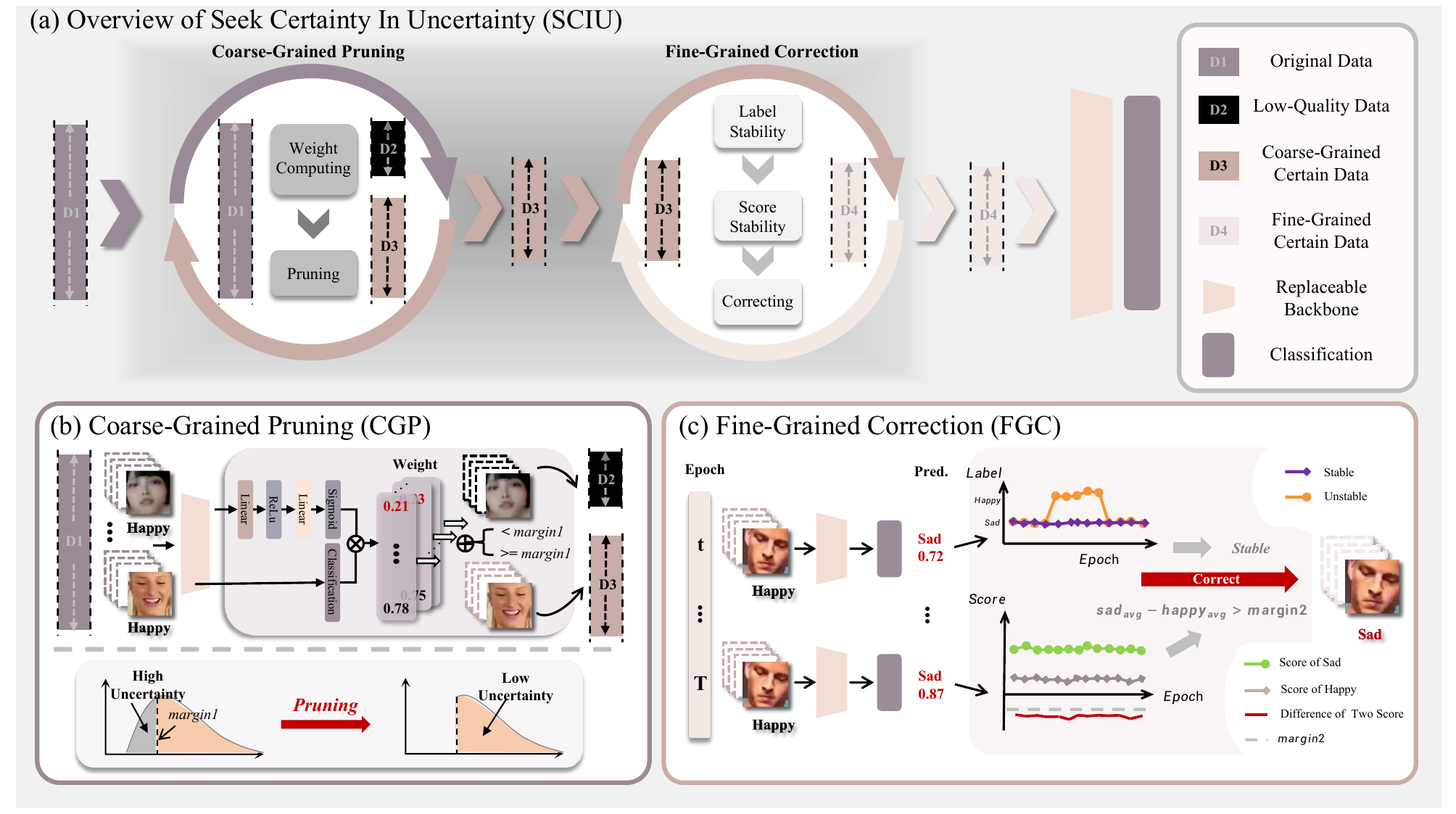}
	\caption{This figure illustrates an overview of our proposed Seeking Certainty in Uncertainty (SCIU) framework. Figure \ref{fig3}(a) outlines the overview of SCIU , comprising two primary stages: Coarse-Grain Pruning (CGP) and Fine-Grain Correction (FGC). The objective of Coarse-Grain Pruning (CGP) is to eliminate the first type of uncertainty by pruning the low-quality and unusable data , resulting in a coarse-grained certain dataset. This subset is then processed in the FGC stage, where the wrong-annotated samples are corrected. Finally, we obtain fine-grained-certain data which are subsequently utilized for model training. Figure \ref{fig3}(b) illustrates that CGP calculates the weight of each sample and prunes those with low weights. Figure \ref{fig3}(c) demonstrates that FGC corrects samples that are stably mispredicted across epochs, thus ensuring correctly annotated samples are used for training.}
    \label{fig3}
\end{figure*}
\vspace{-2mm}

\section{METHOD}
\subsection{Overview}

The SCIU framework, as depicted in Figure \ref{fig3}(a), is architected into two stages: CGP and FGC, which is shown in Figure \ref{fig3}(b) and Figure \ref{fig3}(c). During the CGP stage, we initially estimate the weight of samples within the original DFER dataset and multiply the weight by the logits. Comparing the result with a confidence threshold, we could perform pruning operations. This process yields a dataset with coarse-grained-certain data, which will be used in the FGC stage. Similarly, we predicts sample labels and values, monitors the stability of predictions in two dimension across $n$ rounds in the FGC stage and correct the wrong-annotated samples. Ultimately, we get a dataset with fine-grained-certain data, employing it for subsequent training and continually updating the dataset with accurate and certain data dynamically.

\vspace{-2mm}

\subsection{Coarse-Grained Pruning Stage}

During the CGP stage of our framework, we assign a specific weight to each sample. Denote the dataset as $D_{1} = [v_{1}, v_{2}, \ldots, v_{n}]$. At the $t$-th iteration, we input $v_{i}$ to acquire its embedding $x_{i}$. Subsequently, $x_{i}$ is concurrently directed into two branches as depicted in Figure \ref{fig3}(b). The first path, designated for weight computation, employs a two-layer Multilayer Perceptron (MLP) to compute the weight of $v_{i}$, formulated as follows:
\begin{equation}
w_i = \sigma\left(W_{2}^{T} \text{ReLU}\left(W_{1}^{T} x_{i} + b_{1}\right) + b_{2}\right),
\end{equation}
where $w_i$ signifies the weight for sample $v_{i}$, $x_{i}$ represents the input embedding vector, $W_{1}^{T}$ and $b_{1}$ denote the weight and bias of the initial linear layer respectively, $W_{2}^{T}$ and $b_{2}$ are the weight and bias for the subsequent linear layer, $\sigma$ is the Sigmoid activation function. 



Concurrently with the computation of $w_i$, $x_{i}$ is also fed into the second path which is the classification layer, denoted as $f(x_{i}, \theta, t)$, to obtain the predicted probability value, $p_i = f(x_{i}, \theta, t)$, and here $\theta$ is the learnable parameters. The weighted predicted score of sample $v_i$ during the $t$-th epoch is thus determined by the formula:
\begin{equation}
s_{i}^{t} = w_i \times p_i,
\end{equation}
After getting $s_{i}^{t}$, we will calculate the mean of these scores across the preceding $t$ rounds as follows:
\begin{equation}
S_{T} = \frac{1}{t} \sum_{j=T-t+1}^{T}s_{i}^{j},
\end{equation}
Upon computing $S_{T}$ for $v_i$ in the current $T$-th round, we evaluate it against a predetermined confidence threshold to decide on the necessity for pruning. This decision is formulated as:
\begin{equation}
\alpha = 
\begin{cases} 
1, & \text{if } S_{T} > \lambda, \\
0, & \text{otherwise},
\end{cases}
\end{equation}
where $\alpha$ signifies the pruning requirement. If $\alpha=0$,  pruning is necessary, vice versa. The pruned dataset, $D_{2}$, is formulated as:
\begin{equation}
D_{2} = \{v_{i} \in D | \alpha=0\}^n_{i=1}.
\end{equation}
Consequently, the dataset with coarse-grained-certain data, $D_{3}$, is obtained and formulated as:
\begin{equation}
D_{3} = \{v_{i} \in D | \alpha=1\}^n_{i=1}.
\end{equation}
This process prunes the first type of uncertainty data, which is the low-quality and unusable data, from the original dataset. Thereby preparing a coarse-grained-certain dataset, $D_{3}$, optimized for subsequent FGC stage.
\\ \textbf{Weighted Cross Entropy Loss.}
During both training and inference, we utilize $s_{i}^{t}$ as the novel prediction score, and we initialize the weights during the warm-up phase. We employ a Weighted Cross-Entropy Loss as our loss function, which is formulated as:
\begin{equation}
\mathcal{L}_{\text{WCE}} = \mathcal{L}_{\text{CE}(w_{i}f(x_{i}, \theta),y)},
\end{equation}
which ensures we learn the sample weights and prune away the low-weight samples.
\\ \textbf{Explaining weights and pruning mechanisms from the perspective of gradients.} The model assigns higher weights \( w_i \) to accurately predicted samples because these samples typically have clear and reliable feature representations, resulting in lower loss. Specifically, the predicted values \( f(x_i,\theta) \) for these samples are closer to the true labels \( y \), thereby reducing the error term in the loss function. The gradient of the WCE Loss is calculated as follows:

\[
\frac{\partial \mathcal{L}_{\text{WCE}}}{\partial (w_i f(x_i; \theta))} = -\frac{1}{w_i f(x_i; \theta)} (y - w_i f(x_i; \theta))
\]

where \( y \) represents the true labels. Samples with higher weights have larger gradients, making their influence on the model parameter updates more significant, thereby enhancing the learning of these samples' features. During backpropagation, the model adjusts the weights based on the loss feedback, favoring samples with lower losses by assigning them higher weights, as their predictions are closer to the true labels, providing more reliable supervision signals. The model can autonomously identify and assign weights to data of varying quality. Additionally, we implemented a threshold pruning mechanism to exclude samples with weighted predictions \( w_i f(x_i; \theta) \) below the threshold \( \lambda \). This process aims to further reduce the impact of low-quality samples on model training. In the pruned dataset, only samples with higher weighted predictions are retained, while low-quality samples are assigned lower weights and excluded, minimizing their influence on the model's learning process. Conversely, high-quality samples, due to their higher weights, have a more substantial impact on gradient computation, positively contributing to the model's final performance. And we did some visualization in the Experiment\ref{experiment} part to show that low-quality samples have lower weights than the high-quality samples.

\subsection{Fine-Grained Correction}

Following the CGP stage, we got the coarse-fine-certain dataset $D_{3}$, comprising samples characterized by high-quality and usable for learning. FGC stage aims at rectifying the second variety of uncertainty. In the domain of SSL—illustrated by approaches such as MixMatch \cite{berthelot2019mixmatch} and DivideMix \cite{li2020dividemix}—unlabeled samples are frequently allocated pseudo-labels for the purposes of training, or alternatively, samples bearing high-loss labels are reevaluated and potentially relabeled to facilitate their inclusion in the training process. 
This bears a striking resemblance to FGC of correcting the labels of wrong-annotated samples. The difference lies in the fact that we have already pruned the first type of uncertainty, implying that all the data we use is worth learning from, devoid of what would be considered unusable. The existing uncertain data is of equal value to the certain data: once corrected, it can be utilized for model training.


Our empirical analysis underscores the significance of prediction stability as a determinant for the necessity of label correction. We posit that samples that are consistently predicted as a particular class are potential candidates for mislabeling. Inspired by this insight, we designed FGC stage tailored to eliminate the second type of uncertainty by correcting the wrong-annotated samples.


Let $D_{3} = \{(z_i, y_i) \in (Z, Y)\}$, as illustrated in Figure \ref{fig3}(c). Each sample $z_i$ is input into the backbone, after which we perform classification to derive the predicted label $y'_i$ and the corresponding weighted prediction score $p'_i$. To determine the stability of a sample, we examine it across two dimensions: predicted label stability and prediction score stability.


\noindent \textbf{Predicted Label Stability.}
In the frist dimension, we scrutinize the stability of the predicted label. Specifically, we assess whether the predicted labels remain consistent across the past $t$ rounds, a condition formalized by the equation:
\begin{equation}
\forall t_1, t_2 \in \{t+1, \ldots, T\}, \quad y'^{(t_1)}_{i} = y'^{(t_2)}_{i}
\end{equation}
where $T$ denotes the current epoch.

If the predictions exhibit consistency, it suggests a potential mislabeling of the sample. Nevertheless, this criterion alone does not warrant the correction of the label. Consequently, to refine our assessment and enhance the precision of our correction mechanism, we incorporate an additional evaluation dimension focused on the stability of the predicted probability values.
\noindent \textbf{Prediction Score Stability.} For each sample $z_i$, we define two pivotal metrics to gauge prediction score stability: the average of the predicted probabilities $\bar{p'}$ for the model's predicted label $y'$ across the past $t$ rounds, and the average of the predicted probabilities $\bar{p}_{\text{gt}}$ for the ground truth label $y$ over the same period. These averages are represented by the following formulas:
\begin{equation}
\bar{p'} = \frac{1}{t} \sum_{i=t+1}^{T} p'_i,
\end{equation}

\begin{equation}
\bar{p}_{\text{gt}} = \frac{1}{t} \sum_{i=t+1}^{T} p_{\text{gt}}^{(t)},
\end{equation}

The differential between these two averages, $\Delta p$, quantifies the discrepancy between the model's predictions and the original labels:
\begin{equation}
\Delta p = \bar{p'} - \bar{p}_{\text{gt}},
\end{equation}

If this differential $\Delta p$ surpass a predetermined confidence threshold $\tau$, the predicted pseudo-label is deemed more accurate than the original label, and thus, adopted as the true label. This conditional evaluation is encapsulated as:
\begin{equation}
\beta = 
\begin{cases} 
1, & \text{if } \Delta p > \tau, \\
0, & \text{otherwise}.
\end{cases}
\end{equation}

Here, $\tau$ represents the threshold that delineates the minimum requisite differential between the averaged predicted probabilities for a pseudo-label to be considered as the true label. The decision variable $\beta$ dictates the acceptance of the pseudo-label based on the differential exceeding $\tau$. The resulting fine-grained-certain dataset $D_{4}$, is defined as:
\begin{equation}
D_4 = \{ (z_i, y'_i) \in (Z, Y') \mid \beta = 1 \} \cup \{ (z_i, y_i) \in (Z, Y) \mid \beta = 0 \}^n_{i=1}.
\end{equation}
After obtaining $D_4$, we use it in the final training process.





\section{Experimental Evaluation}

\subsection{Experimental Configuration}
\definecolor{lightgray}{gray}{0.9}
\begin{table*}[!t]
\centering
\caption{Comparison (\%) of our SCIU framework and original framework on six methods (WAR/UAR) on FERV39k, DFEW and MAFW. \textuparrow{} represents a performance improvement compared to the original framework.}
\label{tab1}
\begin{tabular}{cccccccc}
\toprule
& & \multicolumn{2}{c}{FERV39k} & \multicolumn{2}{c}{DFEW} & \multicolumn{2}{c}{MAFW} \\
\cmidrule(lr){3-4} \cmidrule(lr){5-6} \cmidrule(lr){7-8}
Method & Status & WAR & UAR & WAR & UAR & WAR & UAR \\
\midrule
\multirow{2}{*}{Res18\_LSTM \cite{he2016deep,graves2012long}} & Original & 42.59 & 30.92 & 63.85 & 51.32 & 36.49 & 25.83 \\
 & \textbf{SCIU} & \textbf{47.32} (\textbf{4.73\textuparrow}) & \textbf{34.24} (\textbf{3.32\textuparrow}) & \textbf{64.72} (\textbf{0.87\textuparrow}) & \textbf{51.40} (\textbf{0.08\textuparrow}) & \textbf{37.36} (\textbf{0.87\textuparrow}) & \textbf{26.21} (\textbf{0.38\textuparrow}) \\
\midrule
\multirow{2}{*}{Res50\_LSTM \cite{he2016deep,graves2012long}} & Original & 40.75 & 32.12 & 65.53 & 51.83 & 40.51 & 31.18 \\
 & \textbf{SCIU} & \textbf{48.57} (\textbf{7.82\textuparrow}) & \textbf{38.97} (\textbf{6.85\textuparrow}) & \textbf{66.55} (\textbf{1.02\textuparrow}) & \textbf{54.58} (\textbf{2.75\textuparrow}) & \textbf{42.25} (\textbf{1.74\textuparrow}) & \textbf{31.43} (\textbf{0.25\textuparrow}) \\
\midrule
\multirow{2}{*}{VGG13\_LSTM \cite{simonyan2014very,graves2012long}} & Original & 43.37 & 32.41 & 60.62 & 48.36 & 32.41 & 22.04 \\
 & \textbf{SCIU} & \textbf{46.51} (\textbf{3.14\textuparrow}) & \textbf{32.75} (\textbf{0.34\textuparrow}) & \textbf{63.18} (\textbf{2.56\textuparrow}) & \textbf{48.41} (\textbf{0.05\textuparrow}) & \textbf{32.74} (\textbf{0.33\textuparrow}) & \textbf{22.69} (\textbf{0.65\textuparrow}) \\
\midrule
\multirow{2}{*}{VGG16\_LSTM \cite{simonyan2014very,graves2012long}} & Original & 41.7 & 30.93 & 60.23 & 48.03 & 32.51 & 22.32 \\
 & \textbf{SCIU} & \textbf{45.53} (\textbf{3.83\textuparrow}) & \textbf{33.84} (\textbf{2.91\textuparrow}) & \textbf{61.51} (\textbf{1.28\textuparrow}) & \textbf{49.26} (\textbf{1.23\textuparrow}) & \textbf{33.28} (\textbf{0.77\textuparrow}) & \textbf{23.11} (\textbf{0.79\textuparrow}) \\
\midrule
\multirow{2}{*}{Former-DFER \cite{zhao2021former}} & Original & 46.85 & 37.2 & 65.70 & 53.69 & 36.38 & 25.49 \\
 & \textbf{SCIU} & \textbf{49.34} (\textbf{2.49\textuparrow}) & \textbf{38.15} (\textbf{0.95\textuparrow}) & \textbf{67.88} (\textbf{2.18\textuparrow}) & \textbf{53.78} (\textbf{0.09\textuparrow}) & \textbf{37.90} (\textbf{1.52\textuparrow}) & \textbf{26.58} (\textbf{1.09\textuparrow}) \\
\midrule
\multirow{2}{*}{ViT\_LSTM \cite{dosovitskiy2020image,graves2012long}} & Original & 48.67 & 39.77 & 67.07 & 54.75 & 41.6 & 29.62 \\
 & \textbf{SCIU} & \textbf{50.5} (\textbf{1.83\textuparrow}) & \textbf{40.11} (\textbf{0.34\textuparrow}) & \textbf{69.2} (\textbf{2.13\textuparrow}) & \textbf{55.61} (\textbf{0.86\textuparrow}) & \textbf{42.58} (\textbf{0.98\textuparrow}) & \textbf{29.85} (\textbf{0.23\textuparrow}) \\
\midrule
\rowcolor{lightgray}
Average Improvement & & \textbf{3.97\textuparrow} & \textbf{2.45\textuparrow} & \textbf{1.67\textuparrow} & \textbf{0.84\textuparrow} & \textbf{1.20\textuparrow} & \textbf{0.56\textuparrow} \\
\bottomrule
\end{tabular}
\end{table*}

\textbf{Datasets.} The SCIU framework is crafted to mitigate the two types of uncertainty issues in DFER datasets, particularly those encountered in real-world settings. Here we integrates an empirical evaluation using three representative in-the-wild DFER datasets, DFEW \cite{jiang2020dfew}, FERV39k \cite{wang2022ferv39k} and MAFW\cite{liu2022mafw}. The DFEW dataset is an extensive collection of 16,372 video clips extracted from over 1,500 movies, encapsulating a wide array of emotional expressions. In parallel, the FERV39k dataset encompasses 38,935 video clips, methodically segmented into 22 distinct scenes, each posing unique analytical challenges and uncertainty characteristics. MAFW, a large-scale multi-modal compound affective database with 10,045 video-audio clips in the wild. These datasets, notable for their inherent difficulty and the substantial uncertainty levels they exhibit, serve as a rigorous testing ground for the efficacy of the SCIU framework. Implementing our framework across various model architectures has demonstrated significant enhancements in performance metrics across these challenging datasets. More implementation details will be provided in the \textbf{supplementary materials}.

\noindent \textbf{Implementation Details.} The SCIU framework adopts a consistent configuration of 16 frames per video clip across diverse models, adjusting training parameters—including batch size, learning rate, and input size—accordingly, as detailed in Table \ref{tab1}. This uniformity ensures a standardized evaluation basis while accommodating the specific computational demands of each model architecture. Developed on the robust PyTorch platform, our framework and all associated models were trained utilizing the computational prowess of the NVIDIA GeForce RTX 3090 GPUs. The training regimen spanned 200 epochs, leveraging stochastic gradient descent (SGD) with a momentum setting of 0.9 to optimize model performance. To assess the outcomes of our experiments comprehensively, we implemented two critical evaluation metrics: Weighted Average Recall (WAR) and Unweighted Average Recall (UAR).

\subsection{Comparison with Existing Methods}

\textbf{Results on FERV39k.}
Table \ref{tab1} outlines the base performance of various methods extensively applied in DFER tasks, as well as their performance improvements after integrating the SCIU framework. The analysis encompassed six widely used methods in DFER studies: ResNet18\_LSTM  \cite{he2016deep,graves2012long}, ResNet50\_LSTM  \cite{he2016deep,graves2012long}, VGG13\_LSTM  \cite{simonyan2014very,graves2012long}, VGG16\_LSTM  \cite{simonyan2014very,graves2012long}, Former-DFER  \cite{zhao2021former}, and ViT\_LSTM  \cite{dosovitskiy2020image}. To accommodate our framework, each model was adapted by adding an extra branch after the embedding layer for weight assessment, without altering other components. The framework notably excelled in mitigating uncertainty challenges within FERV39k, demonstrating an average uplift of 3.97\% in WAR and 2.45\% in UAR. Particularly, the ResNet50\_LSTM model showed a remarkable improvement of \textbf{7.82\%} in WAR and \textbf{6.85\%} in UAR.

\noindent \textbf{Results on DFEW.}
The performance outcomes for the six methods on the DFEW dataset are detailed in Table \ref{tab1}. While the enhancements on FERV39k were particularly striking, DFEW also witnessed notable improvements, with an average increase of 1.67\% in WAR and 0.84\% in UAR. Although these gains are slightly more modest compared to those observed on FERV39k, they represent significant advancements within the realm of DFER tasks. The differential impact can be attributed to that DFEW has fewer annotation errors and a lower error rate, as evidenced by the fact that many methods achieve higher accuracy on this smaller dataset.

\noindent \textbf{Results on MAFW.}
Table \ref{tab1} also presents the performance of the six methods on the new dataset. The integration of the SCIU framework yielded notable improvements, albeit to a lesser extent compared to FERV39k and DFEW. The average enhancement was 1.20\% in WAR and 0.56\% in UAR. Despite being more modest, these improvements are still significant within the context of DFER tasks. The SCIU framework's effectiveness on MAFW indicates its robustness across different datasets.

\noindent \textbf{Further Analysis on Experiment Results.}
Compared to DFEW, FERV39k exhibits increased complexity and a wider array of scenes, resulting in intensified problems related to both categories of uncertainty. The rich scene diversity in FERV39k presents significant challenges in annotation, resulting in a more extensive range of noise complexities. This complexity is a major factor behind the marked performance improvements achieved using the SCIU framework on FERV39k, in contrast to the enhancements observed on DFEW. This differential impact is reflected in the extent of data filtered and corrected by SCIU: FERV39k saw 11,707 samples pruned and 1,436 corrected, while DFEW had approximately 2,272 pruned and 974 corrected, which is shown in Table \ref{tab7}. Despite the proportional data differences not being stark, the voluminous data pool of FERV39k underscores the effectiveness of the pruning and correction of SCIU, evidencing over 2.30\% WAR and 1.61\% UAR on average improvement comparing with SCIU on DFEW.
\begin{figure*}[htbp]
\centering
	\includegraphics[width=\textwidth]{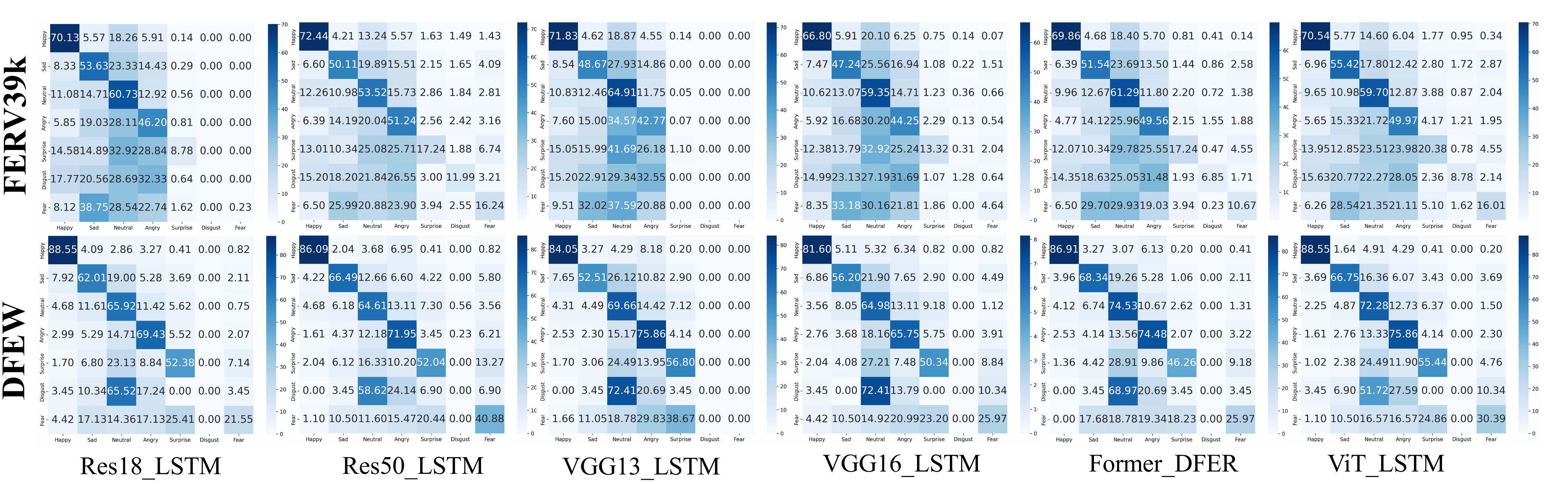}
	\caption{Confusion Matrix of different methods on FERV39k and DFEW.}
    \label{fig5}
\end{figure*}
Moreover, confusion matrices for the six methods across both DFEW and FERV39k datasets are presented, offering empirical evidence of consistent performance of SCIU. These matrices, showcased in Figure \ref{fig5}, further attest to the adaptability and efficacy of SCIU in enhancing recognition accuracy across a variety of methods and datasets, reinforcing the broad applicability and effectiveness of SCIU in addressing uncertainty issues in DFER dataset.

\vspace{-3mm}
\subsection{Ablation Study}

In this section, ablation studies are conducted using ResNet18\_LSTM \cite{he2016deep,graves2012long} as the foundational model on two significant datasets, DFEW and FERV39k. These studies focus on evaluating the individual and combined impacts of the CGP and FGC, alongside the examination of hyperparameter variations. 


\noindent \textbf{Effectiveness of Dual-Stage Mechanism of SCIU.} In this paper, we focus on the the dual-stage of SCIU. The influence of each stage was meticulously assessed on the DFEW and FERV39k datasets, as documented in Tables \ref{tab3} and \ref{tab7}. Within the FERV39k training set, consisting of 31,088 samples, the application of solely the CGP stage resulted in the retention of 17,257 samples by the final training epoch—indicating a filtration of approximately 44.49\% (13,831 samples). This process, which utilized about half of the dataset, led to an enhancement of 3.54\% in WAR and 2.09\% in UAR. Subsequently, employing only FGC resulted in the correction of 8,271 samples by the last epoch, yielding improvements of 2.51\% in WAR and 0.71\% in UAR. Nevertheless, the integration of both stages culminated in the filtration of 11,707 samples and correction of 1,436 samples, significantly boosting performance by 4.73\% in WAR and 3.32\% in UAR, thereby underscoring the comprehensive efficacy of the SCIU framework. For the DFEW dataset, the isolated application of either CGP or FGC minimally impacted WAR and UAR, with pruning and correction numbers standing at 2,103 and 2,064, respectively. However, the synergistic application of both stages markedly amplified the WAR by 0.87\% and UAR by 0.08\%, with the pruning and correction numbers adjusting to 851 and 1,168. 

This corroborates the presence of the two types of uncertainty issues we proposed. The CGP adeptly addresses the first type of uncertainty issue by removing low-quality, unusable samples, while the FGC resolves the second type, correcting wrong-annotated samples. However, addressing uncertainty in DFER dataset with just one approach does not yield optimal results; when both CGP and FGC are employed concurrently, the model demonstrates enhanced performance.

\setlength{\tabcolsep}{1.5pt}
\begin{table}[t]
\centering
\caption{Ablation study (\%) with different stages in SCIU. \textuparrow{ }represents improvement of performance compared to not using the two stages. }
\label{tab3}
\begin{tabular}{cccccc}
\toprule
\multirow{2}{*}{CGP} & \multirow{2}{*}{FGC} & \multicolumn{2}{c}{FERV39k} & \multicolumn{2}{c}{DFEW} \\
\cmidrule(lr){3-4} \cmidrule(lr){5-6}
& & WAR & UAR & WAR & UAR \\
\midrule
\ding{55} & \ding{55} & 42.59 & 30.92 & 63.85 & 51.32 \\
\ding{51} & \ding{55} & 46.13 (3.54\textuparrow) & 33.01 (2.09\textuparrow) & 64.25 (0.40\textuparrow) & 50.75 (0.57\textdownarrow) \\
\ding{55} & \ding{51} & 45.1 (2.51\textuparrow) & 31.63 (0.71\textuparrow) & 64.42 (0.57\textuparrow) & 51.23 (0.09\textdownarrow) \\
\rowcolor{gray!30} 
\ding{51} & \ding{51} & \textbf{47.32} 
 (4.73\textuparrow) & \textbf{34.24} (3.32\textuparrow) & \textbf{64.72} (0.87\textuparrow) & \textbf{51.40} (0.08\textuparrow) \\ 
\bottomrule
\end{tabular}
\end{table}

\setlength{\tabcolsep}{5pt}
\begin{table}[htbp]
\centering
\caption{Ablation study with different stages effection on the datasets. Pruning represents the pruned sample number of CGP, and Correction represents the corrected sample number.}
\label{tab7}
\begin{tabular}{cccccc}
\toprule
\multirow{2}{*}{CGP} & \multirow{2}{*}{FGC} & \multicolumn{2}{c}{FERV39k} & \multicolumn{2}{c}{DFEW} \\
\cmidrule(lr){3-4} \cmidrule(lr){5-6}
 &  & Pruning & Correction & Pruning & Correction \\
\midrule
\ding{51} & \ding{55} & 13,831 & \diagbox[dir=NE,innerwidth=0.3cm]{}{} & 2,103 & \diagbox[dir=NE,innerwidth=0.3cm]{}{} \\
\ding{55} & \ding{51} & \diagbox[dir=NE,innerwidth=0.3cm]{}{} & 8,271 & \diagbox[dir=NE,innerwidth=0.3cm]{}{} & 2,064 \\

\ding{51} & \ding{51} & 11,707 & 1,436 & 851 & 1,168 \\
\bottomrule
\end{tabular}
\end{table}

\setlength{\tabcolsep}{2.2pt} 
\begin{table}[ht]
\centering
\caption{Ablation study (\%) of different $\lambda$ and $\tau$ values on FERV39k and DFEW.}
\label{tab4}
\begin{subtable}{.5\linewidth}
\caption{CGP threshold \(\lambda\)}
\centering
\begin{tabular}{ccccc}
\toprule
\multirow{2}{*}{\(\lambda\)} & \multicolumn{2}{c}{FERV39k} & \multicolumn{2}{c}{DFEW} \\
& WAR & UAR & WAR & UAR \\
\midrule
0.5 & 46.85 & 33.92 & 64.29 & 50.39 \\
0.6 & 46.9 & 34.45 &\textbf{ 64.72} & \textbf{51.40} \\
0.7 &\textbf{ 47.32} & \textbf{34.24} & 64.03 & 50.90 \\
0.8 & 46.32 & 33.15 & 62.37 & 47.81 \\
0.9 & 45.75 & 32.09 & 61.77 & 47.65 \\
\bottomrule
\end{tabular}
\end{subtable}%
\begin{subtable}{.5\linewidth}
\centering
\caption{FGC threshold \(\tau\)}
\begin{tabular}{ccccc}
\toprule
\multirow{2}{*}{\(\tau\)} & \multicolumn{2}{c}{FERV39k} & \multicolumn{2}{c}{DFEW} \\
& WAR & UAR & WAR & UAR \\
\midrule
0.1 & 46.32 & 33.65 & 63.05 & 50.41 \\
0.2 & \textbf{47.32} & \textbf{34.24} & 63.39 & 50.64 \\
0.3 & 46.83 & 33.52 & 63.48 & 50.77 \\
0.4 & 46.34 & 33.16 & \textbf{64.72} &\textbf{ 51.40} \\
0.5 & 46.13 & 33.54 & 64.03 & 50.90 \\
\bottomrule
\end{tabular}
\end{subtable}
\end{table}

\setlength{\tabcolsep}{8pt}
\begin{table}{}
\centering
\caption{Ablation study (\%) of different cumulative epoches ($t$) on FERV39k and DFEW.}
\label{tab5}
\begin{tabular}{ccccc}
\toprule
\multirow{2}{*}{\(t\)} & \multicolumn{2}{c}{FERV39k} & \multicolumn{2}{c}{DFEW} \\
& WAR & UAR & WAR & UAR \\
\midrule
2 & 47.04 & 33.86 & \textbf{64.72} &\textbf{ 51.40} \\
3 & \textbf{47.32} & \textbf{34.24} & 62.2 & 48.65 \\
4 & 46.86 & 33.79 & 62.71 & 48.13 \\
5 & 46.72 & 33.76 & 61.34 & 48.96 \\
\bottomrule
\end{tabular}
\end{table}
\begin{figure*}[t]
\centering
	\includegraphics[width=1\textwidth]{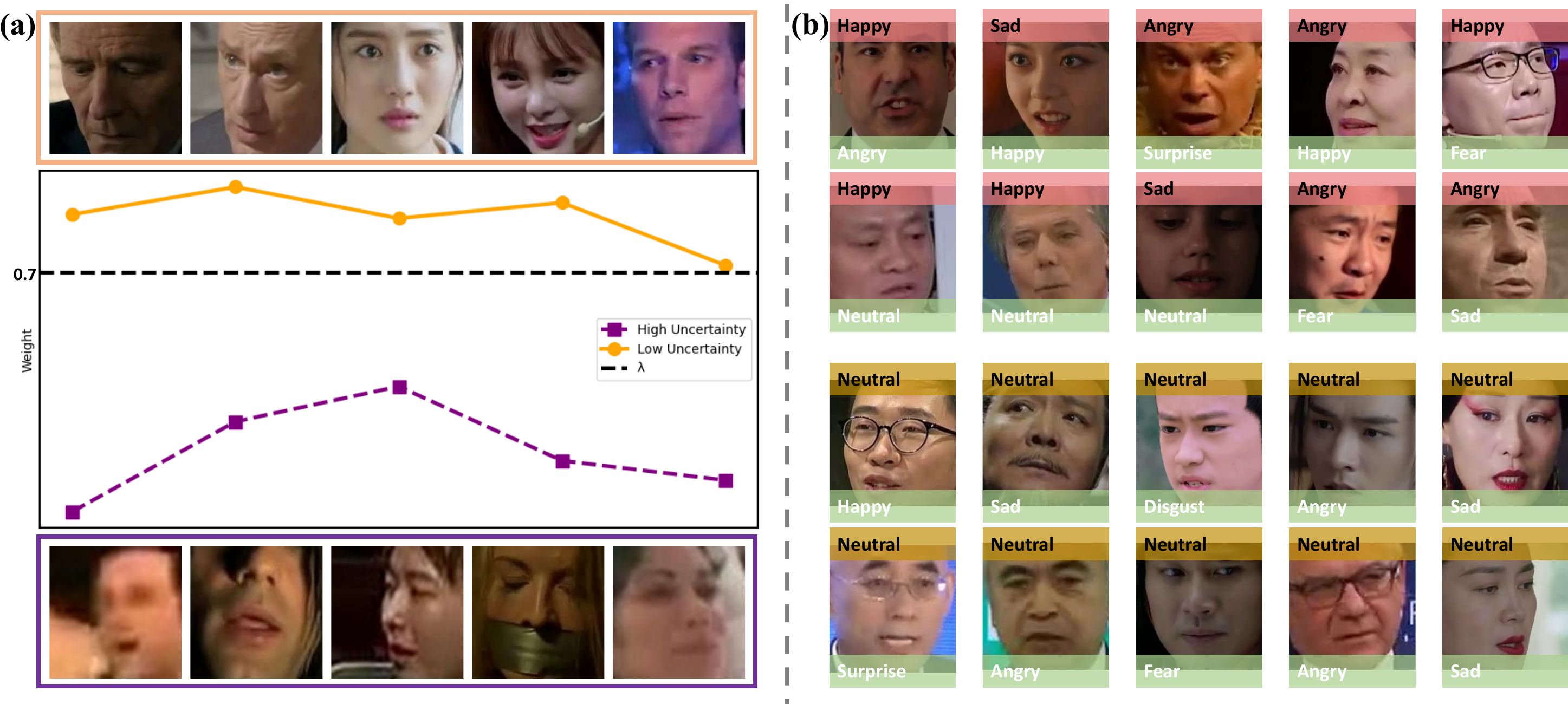}
	\caption{Illustration of the pruning and correcting samples in FERV39k and DFEW. (a) illustrate the weight distribution of selected samples (high quality) and pruned samples (low quality). (b) illustrates the corrected samples. Upper group is the non-neutral to corrected label, and lower group is the 'Neutral' to corrected label.}
    \label{fig6}
\end{figure*}

\noindent \textbf{Impact of Hyperparameter.}
Here we utilizing ResNet18\_LSTM as a representative model, we employed a detailed exploration to ascertain the influence of three critical hyperparameters on model performance. This examination was specifically tailored to the difference of the DFEW and FERV39k datasets, with anticipations of distinct outcomes and hyperparameter optimizations for each. The forthcoming section is dedicated to an in-depth analysis of these variations. The most optimal hyperparameters derived from our experimental analysis across various methods, highlighting the tailored configurations necessary for optimizing the SCIU framework's performance with different datasets and architectures, which is provided in the \textbf{supplementary materials}.
\\ \textbf{Evaluation of $\lambda$ and $\tau$.} A primary focus was placed on $\lambda$, exploring a range between 0.5 to 0.9, with the findings documented in Table \ref{tab4}(a). The investigation uncovered that excessively high values of $\lambda$ lead to the removal of a substantial portion of the dataset, potentially undermining the ability of model to fit. Conversely, overly low values of 
$\lambda$ may not sufficiently prune out the first type of certainty, inadvertently retaining undesirable data elements. Significantly, the optimal 
$\lambda$ value displayed minor discrepancies, which FERV39k is 0.7 and DFEW is 0.6, highlighting the importance of tailoring hyperparameter adjustments to the specific dataset. Similarly, our investigation into $\tau$, paralleled the insights gained from $\lambda$. This analysis, summarized in Table \ref{tab4}(a), showed that an excessively high $\tau$ reduces the correction number, potentially neglecting the mislabeled samples. Conversely, a very low $\tau$ might lead to overcorrection, risking the introduction of additional noise. After examining values from 0.1 to 0.4, we identified 0.2 as the optimal $\tau$ for FERV39k, and 0.4 for DFEW.

\noindent \textbf{Evaluation of $t$.} The evaluation period for determining data stability, a key factor in our analysis, was also scrutinized. While longer periods might intuitively seem to ensure greater reliability in data consistency, our experiments, detailed in Table \ref{tab5}, indicate that a duration of 3 epochs for FERV39k and 2 epochs for DFEW is most effective. This finding suggests that overly prolonged assessments of stability might restrict the learning potential, leading it to overfit to certain features or samples and neglect broader learning opportunities. More confusion matrices of ablation study will be provided in the \textbf{supplementary materials}.

\subsection{Visualization}

Visual representations of the samples processed through the two stages of our SCIU framework are depicted in Figure \ref{fig6}. In Figure \ref{fig6}(a), we showcase the weights assigned to samples that were pruned out. These samples exhibit notably low weights. In Figure \ref{fig6}(b), we illustrate some of the corrected samples. The upper group illustrates adjustments from non-neutral to corrected labels, while the lower group shows changes from neutral to corrected labels. This visual evidence clearly demonstrates the enhanced reliability of the pseudo labels FGC corrected, underscoring the capacity SCIU to eliminate the two types of uncertainty within the dataset effectively.


\vspace{-3mm}
\section{Conclusion}

This work introduces the SCIU, a plug-and-play and general framework designed to eliminate the two types of uncertainty issues identified in DFER datasets. The framework creatively segregates the process into two main stages: 1) CGP, which identifies and prunes data characterized as low-quality and unusable, 2) FGC, aiming to correct wrongly annotated data. In this context, the analysis reveals two kinds of incorrect annotations: confirmation bias among annotators and the mislabelling of low-emotion-intensity samples as 'Neutral'.

We deployed our SCIU framework across multiple common methods within DFER tasks, demonstrating the universality of the SCIU framework. It achieved significant performance improvements on DFEW, FERV39k and MAFW datasets, particularly on FERV39k. This not only proves the strength of our method but also confirms the presence of the two types of uncertainties we proposed in DFER datasets. We believe that future DFER methodologies could greatly enhance the foundational performance of model by initially applying SCIU to process DFER datasets before proceeding with method design.

\bibliographystyle{ACM-Reference-Format}
\bibliography{sample-base}

\end{document}